\documentclass[runningheads]{llncs}
\usepackage{graphicx}

\usepackage{tikz}
\usepackage{comment}
\usepackage{amsmath,amssymb}
\usepackage{color}
\usepackage{enumitem}
\usepackage{xcolor}
\usepackage{multirow}
\usepackage{verbatim}
\usepackage{diagbox}
\usepackage{amssymb}
\usepackage{pifont}
\usepackage{cite}
\usepackage{colortbl}
\usepackage{hyperref}
\hypersetup{colorlinks=true,
            linkcolor=red,
            anchorcolor=blue,
            citecolor=green}

\usepackage[accsupp]{axessibility}  

\begin{document}
\pagestyle{headings}
\mainmatter
\def\ECCVSubNumber{2373}  

\title{Learning Dynamic Facial Radiance Fields for Few-Shot Talking Head Synthesis} 



\titlerunning{Learning DFRF for Few-Shot Talking Head Synthesis}
%
\author{Shuai Shen\inst{1,2} \and Wanhua Li\inst{1,2}\and Zheng Zhu\inst{3}\and Yueqi Duan\inst{4} \and Jie Zhou\inst{1,2} \and Jiwen Lu\inst{1,2,}$^{\ast}$}

\authorrunning{Shen et al.}

\institute{Department of Automation, Tsinghua University, China \and Beijing National Research Center for Information Science and Technology, China \and PhiGent Robotics\and
Department of Electronic Engineering, Tsinghua University, China\\
\email{shens19@mails.tsinghua.edu.cn; li-wh17@tsinghua.org.cn; zhengzhu@ieee.org; \{duanyueqi, jzhou, lujiwen\}@tsinghua.edu.cn}}

\maketitle
\let\thefootnote\relax\footnotetext{$^{\ast}$Corresponding author}

\begin{abstract}
Talking head synthesis is an emerging technology with wide applications in film dubbing, virtual avatars and online education. Recent NeRF-based methods generate more natural talking videos, as they better capture the 3D structural information of faces. However, a specific model needs to be trained for each identity with a large dataset. In this paper, we propose Dynamic Facial Radiance Fields (DFRF) for few-shot talking head synthesis, which can rapidly generalize to an unseen identity with few training data. Different from the existing NeRF-based methods which directly encode the 3D geometry and appearance of a specific person into the network, our DFRF conditions face radiance field on 2D appearance images to learn the face prior. Thus the facial radiance field can be flexibly adjusted to the new identity with few reference images. Additionally, for better modeling of the facial deformations, we propose a differentiable face warping module conditioned on audio signals to deform all reference images to the query space. Extensive experiments show that with only tens of seconds of training clip available, our proposed DFRF can synthesize natural and high-quality audio-driven talking head videos for novel identities with only 40k iterations. We highly recommend readers view our supplementary video for intuitive comparisons. Code is available in \url{https://sstzal.github.io/DFRF/}.
\keywords{few-shot talking head synthesis, neural radiance fields}
\end{abstract}

\section{Introduction}

Audio-driven talking head synthesis is an ongoing research topic with a variety of applications including filmmaking, virtual avatars, video conferencing and online education~\cite{zakharov2019few,guo2021ad,zhang2020davd,wang2021one,chen2020talking,zhou2020makelttalk}. Existing talking head generation methods can be roughly divided into 2D-based and 3D-based ones. Conventional 2D-based methods usually depend on GAN model~\cite{das2020speech,gu2020flnet,christos2020headgan} or image-to-image translation~\cite{zhang2020davd,zhou2020makelttalk,eskimez2021speech,zhou2019talking}. However, due to the lack of 3D structure modeling, most of these approaches struggle in generating vivid and natural talking styles. Another genre for talking head synthesis~\cite{linsen2020ebt,yi2020audio,thies2020neural,chen2020talking} relies on the 3D morphable face model (3DMM)~\cite{zollhofer2018state,blanz1999morphable,thies2016face2face}. Benefit from the 3D-aware modeling, they can generate more vivid talking faces than 2D-based methods. Since the use of intermediate 3DMM parameters leads to some information loss, the audio-lip consistency of the generated videos may be affected~\cite{guo2021ad}. 

\begin{figure}[tb]
\begin{center}
\includegraphics[width=1\linewidth]{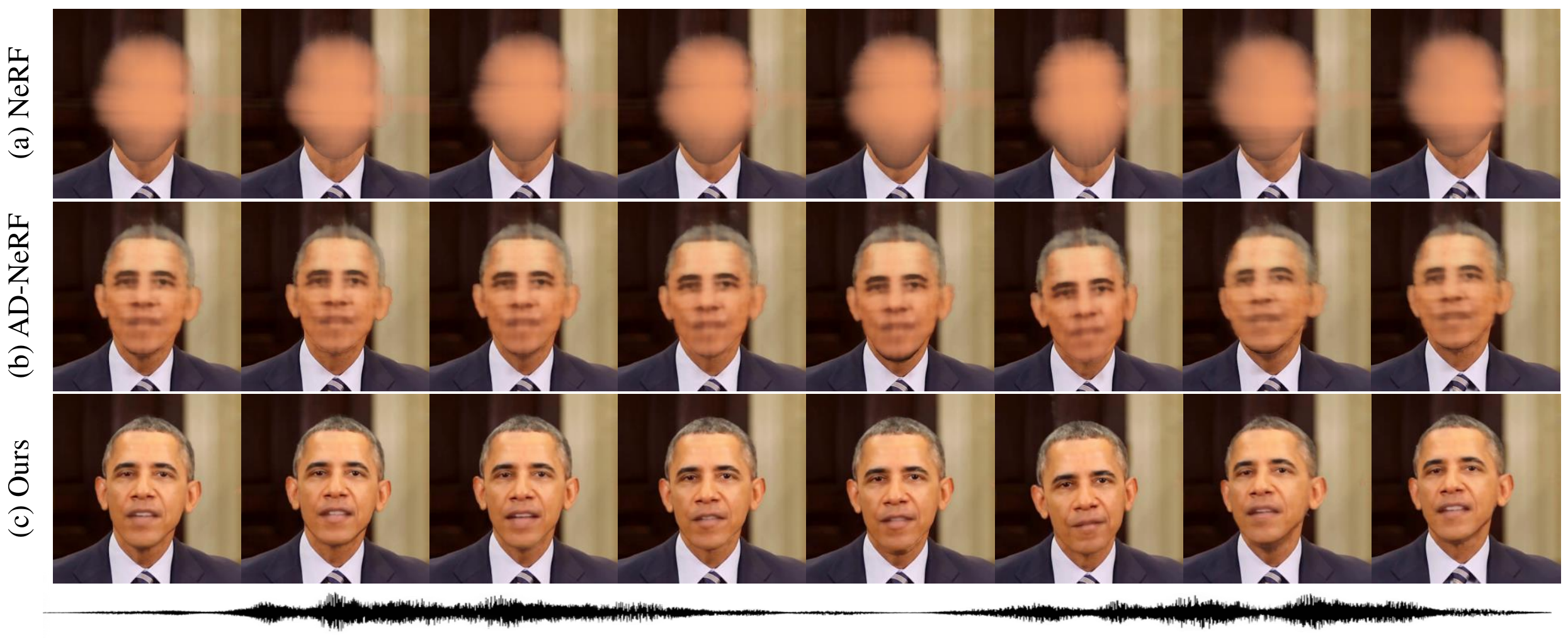}
\end{center}
\caption{We propose Dynamic Facial Radiance Fields (DFRF), a learning framework for few-shot talking head synthesis within a small number of training iterations. Given only a 15s video clip of Obama for 10k iterations training, our DFRF rapidly generalizes to this specific identity including the scene, and synthesizes photo-realistic talking head sequence as shown in row (c). In contrast, NeRF~\cite{mildenhall2020nerf} and AD-NeRF~\cite{guo2021ad} fail to produce plausible results in such a few-shot setting within limited training iterations.}
\label{first}
\end{figure}

More recently, the emerging Neural Radiance Fields (NeRF) based talking head methods~\cite{mildenhall2020nerf,guo2021ad,yao2022dfa} have achieved great performance improvement. They map audio features to a dynamic radiance field for talking portraits rendering without introducing extra intermediate representation. However, they directly encode the 3D geometry and appearance of a specific person into the radiance field, thereby failing to generalize to novel identities. A specific model needs to be trained for each novel identity with high computational cost. Moreover, a large training dataset is required, which cannot meet some practical scenarios where only a few data is available. As shown in Fig.~\ref{first}, given only a 15s training clip, AD-NeRF~\cite{guo2021ad} renders some blurry faces after 10k training iterations.

In this paper, we study this more challenging setting, few-shot talking head synthesis, for the aforementioned practical application scenarios. For an arbitrary new identity with merely a short training video clip available, the model should generalize to this specific person within a few iterations of fine-tuning. There are three key features of the few-shot talking head synthesis \emph{i.e.}\ limited training video, fast convergence, and realistic generation results. To this end, we propose a Dynamic audio-driven Facial Radiance Field (DFRF) for few-shot talking head synthesis. A reference mechanism is designed to learn the generic mapping from a few observed frames to the talking face with corresponding appearance (including the same identity, hairstyle and makeup). Specifically, with some 2D observations as references, the 3D query point can be projected back to the 2D image space of these references respectively and draw the corresponding pixel information to guide the following synthesis and rendering. A prior assumption for such projection operation is that two intersecting rays in 3D volume space should correspond to the same color~\cite{park2020deformable,mildenhall2020nerf}. This conception holds for static scenes, yet talking heads are deformable objects and such naive warping may lead to some mismatch. We therefore introduce a differentiable face warping module for better modeling the facial dynamics when talking. This face warping module is realized as a 3D point-wise deformation field conditioned on audio signals to warp all reference images to the query space.

Extensive experiments show that our proposed DFRF can generate realistic and natural talking head videos with few training data and training iterations. Fig.~\ref{first} shows the visual comparison with NeRF~\cite{mildenhall2020nerf} and AD-NeRF~\cite{guo2021ad}. Given only a 15-second video clip of Obama for 10k training iterations, our proposed DFRF quickly generalizes to this specific identity and synthesizes photo-realistic talking head results. In contrast, NeRF and AD-NeRF fail to produce plausible results in such few-shot setting within limited training iterations. To summarize, we make the following contributions:
\begin{itemize}[leftmargin=*]
\item We propose a dynamic facial radiance field conditioned on the 3D aware reference image features. The facial field can rapidly generalize to novel identities with only 15s clip for fine-tuning.
\item For better modeling the face dynamics of talking head, we learn a 3D point-wise face warping module conditioned on audio signals for each reference image to warp it to the query space.
\item The proposed DFRF can generate vivid and natural talking head videos using only a handful of training data with limited iterations, which far surpasses other NeRF-based methods under the same setting. We highly recommend readers view the supplementary videos for better comparisons.
\end{itemize}

\section{Related Work}

\textbf{2D-Based Talking-Head Synthesis.} 
Talking-head synthesis aims to animate portraits with given audios. 2D-based methods usually employ GANs~\cite{prajwal2020lip,das2020speech,gu2020flnet,christos2020headgan} or image-to-image translation~\cite{zhang2020davd,zhou2020makelttalk,eskimez2021speech,zhou2019talking,zhu2018arbitrary} as the core technologies, and use some intermediate parameters such as 2D landmarks~\cite{chen2019hierarchical,das2020speech,zhou2020makelttalk,chung2017you,lu2021live} to realize the synthesis task. There are also some works focusing on the few-shot talking head generation~\cite{zakharov2019few,wang2021audio2head,meshry2021learned,eskimez2021speech,kumar2020robust}. Zakharov~\emph{et al.}~\cite{zakharov2019few} propose a few-shot adversarial learning approach through pre-training high-capacity generator and discriminator via meta-learning. Wang~\emph{et al.}~\cite{wang2021audio2head} realize one-shot talking head generation by predicting flow-based motion fields. Meshry~\emph{et al.}~\cite{meshry2021learned} disentangle the spatial and style information for few-shot talking head synthesis. However, since these 2D-based methods cannot grasp the 3D structure of head, the naturalness and vividness of the generated talking videos are inferior to the 3D-based methods.

\textbf{3D-Based Talking-Head Synthesis.} A series of 3D model-based methods~\cite{karras2017audio,cudeiro2019capture,fried2019text,ji2021audio,suwajanakorn2017synthesizing,chen2020talking,thies2020neural,linsen2020ebt,jaderberg2015spatial} generate talking heads by utilizing 3D Morphable Models (3DMM)~\cite{zollhofer2018state,blanz1999morphable,thies2016face2face,shang2020self}. Taking advantage of 3D structure modeling, these approaches can achieve more natural talking style than 2D methods. Representative methods ~\cite{thies2020neural,suwajanakorn2017synthesizing} have generated realistic and natural talking head videos. However, since their networks are optimized on a specific identity for idiosyncrasies learning, per-identity training on a large dataset is needed. Another common limitation is the information loss brought by the use of intermediate 3DMM parameters~\cite{blanz1999morphable}. In contrast, our proposed method gets rid of such computationally expensive per-identity training settings while generating high-quality videos. More recently, the emerging NeRF~\cite{mildenhall2020nerf} provides a new technique for 3D-aware talking head synthesis. Guo~\emph{et al.}~\cite{guo2021ad} are the first to apply NeRF into the area of talking head synthesis and have achieved better visual quality. Yao~\emph{et al.}~\cite{yao2022dfa} further disentangle lip movements and personalized attributes. However both of them suffer in the few-shot learning setting.

\textbf{Neural Radiance Fields.} Neural Radiance Fields (NeRF)~\cite{mildenhall2020nerf} store the information of 3D geometry and appearance in terms of voxel grids~\cite{tewari2020state,sitzmann2019scene} with a fully-connected network. The invention of this technology has inspired a series of following works. pi-GAN~\cite{chan2021pi} proposes a generative model with NeRF as the backbone for static face generation while our method learns a dynamic radiance field. Since the original NeRF is designed for static scenes, some works try to extend this technique to the dynamic domain~\cite{park2020deformable,gafni2021dynamic,tretschk2021non,pumarola2021d,gao2020portrait}. Gafni~\emph{et al.}~\cite{gafni2021dynamic} encode the expression parameters into the NeRF for dynamic faces rendering. \cite{park2020deformable,pumarola2021d,tretschk2021non} encode non-rigid scenes via ray bending into a canonical space. ~\cite{wang2021one} represents face as compact 3D keypoints and performs keypoint driven animation. i3DMM~\cite{yenamandra2021i3dmm} generates faces relying on geometry latent code. However, these methods need to optimize the model to every scene independently requiring a large dataset, while our method realizes fast generalization across identities based on easily accessible 2D reference images. There are also some other works that try to improve NeRF's generalization capabilities~\cite{trevithick2021grf,wang2021ibrnet,yu2021pixelnerf}, yet their research are limited to static scenes.

\section{Methodology}
\subsection{Problem Statement} Some limitations of existing talking head technologies hinder them from practical applications. 2D-based methods struggle to generate a natural talking style~\cite{thies2020neural}. Classical 3D-based approaches have information loss due to the use of 3DMM intermediate representations~\cite{guo2021ad}. NeRF-based ones synthesize superior talking head videos, however the computational cost is relatively high since a specific model needs to be trained for each identity. And a large dataset is required for training. We therefore focus on a more challenging setting for the talking head synthesis task. For an arbitrary person with merely a short training video clip available, a personalized audio-driven portrait animation model with high-quality synthesis results should be constructed within only a few iterations of fine-tuning. Three core features of this setting can be summarized as: limited training data, fast convergence and excellent generation effect. 

To this end, we propose a Dynamic Facial Radiance Field (DFRF) for few-shot talking head synthesis. The image features are introduced as a condition to build a fast mapping from reference images to the corresponding facial radiance field. For better modeling the facial deformations, we further design a differentiable face warping module to warp reference images to the query space. Specifically, for fast convergence, a base model is firstly trained across different identities to capture the structure information of the head and establish a generic mapping from audio to lip motions. On this basis, efficient fine-tuning is performed to quickly generalize to a new target identity. In the following, we will detail these designs.

\begin{figure}[tb]
\begin{center}
\includegraphics[width=1\linewidth]{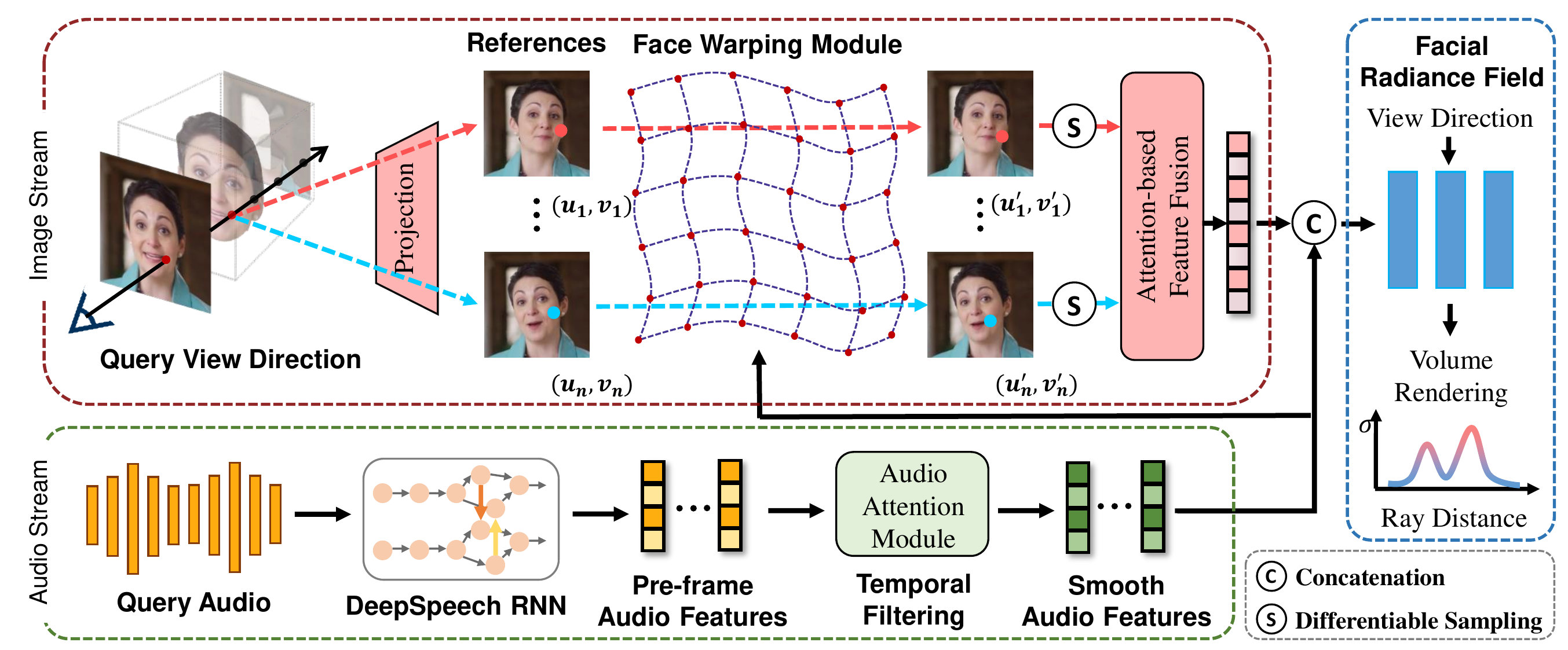}
\end{center}
\caption{Overview of the proposed Dynamic Facial Radiance Fields (DFRF).}
\label{overview}
\end{figure}

\subsection{Dynamic Facial Radiance Field}
\label{3_2}
The emerging NeRFs~\cite{mildenhall2020nerf} provide a powerful and elegant framework for 3D scene representation. It encodes a scene into a 3D volume space with a MLP $\mathcal{F}_{\theta }$. The 3D volume can then be rendered into images by integrating colors and densities along camera rays~\cite{niemeyer2020differentiable,curless1996volumetric,seitz1999photorealistic}. Specifically, using $\mathcal{P}$ as the collection of all 3D points in the voxel space, with a 3D query point ${p} = (x, y, z)\in \mathcal{P}$ and a 2D view direction ${d} = (\theta, \phi)$ as input, this MLP infers the corresponding RGB color $c$ and density $\sigma$, which can be formulated as $\left ( c, \sigma  \right ) = \mathcal{F}_{\theta }\left ( {p}, {d} \right )$.

In this work, we employ NeRF as the backbone for 3D-aware talking head modeling. The talking head task focuses on the audio-driven face animation. However, the original NeRF is designed for only static scenes. We therefore provide the missing deformation channel by introducing audio condition as shown in the audio stream of Fig.~\ref{overview}. We firstly use a pre-trained RNN-based DeepSpeech~\cite{hannun2014deep} module to extract the per-frame audio feature. For inter-frame consistency, a temporal filtering module~\cite{thies2020neural} is further introduced to compute smooth audio features $A$, which can be denoted as the self-attention-based fusion of its neighbor audio features. Taking these audio feature sequences $A$ as the condition, we can learn the audio-lip mapping. This audio-driven facial radiance field can be denoted as $\left ( c, \sigma  \right ) = \mathcal{F}_{\theta }\left ( {p}, {d}, {A} \right )$.

\begin{figure}[tb]
\begin{center}
\includegraphics[width=0.65\linewidth]{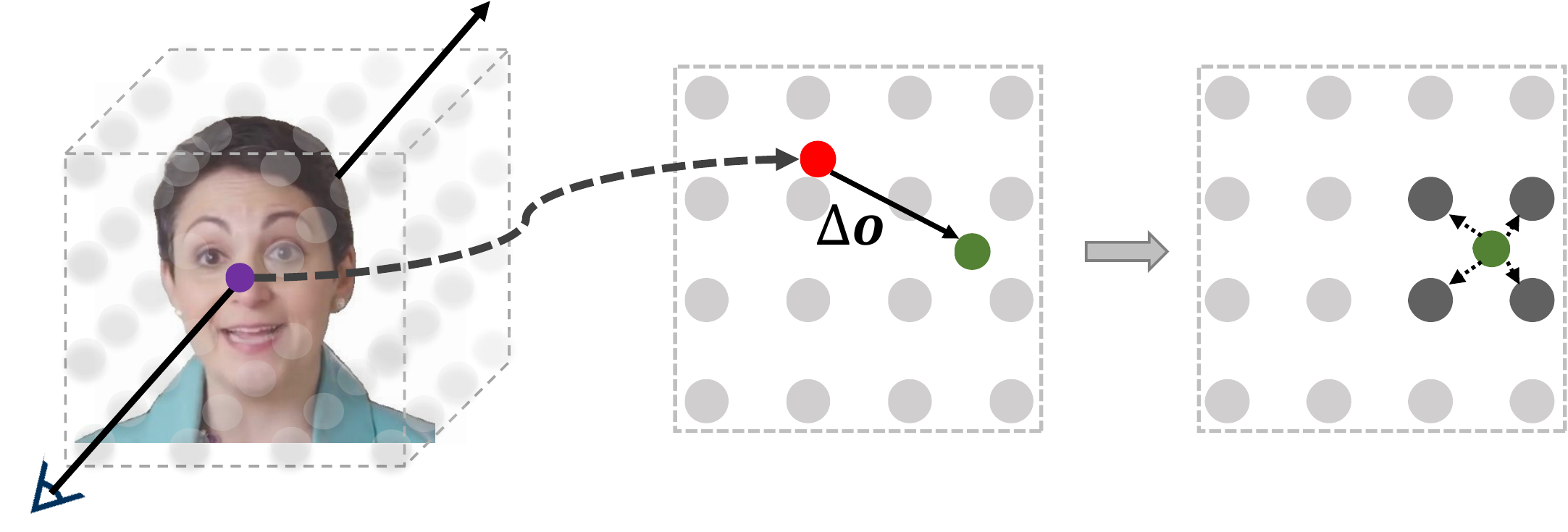}
\end{center}
\caption{Visualization of the differentiable face warping. A query 3D point (purple) is projected to the reference image space (red). Then an offset $\Delta o$ is learned to warp it to the query space (green), where its feature is computed by bilinear interpolation.}
\label{warp}
\end{figure}

Since the identity information is implicitly encoded into the facial radiance field, and no explicit identity feature is provided when rendering, this facial radiance field is person specific. For each new identity, it needs to be optimized from scratch on a large dataset. This leads to expensive calculation costs and requires long training videos. To get rid of these restrictions, we design a reference mechanism to empower a well-trained base model to quickly generalize to new person categories, with only a short clip of the target person available. An overview of this reference-based architecture is shown in Fig.~\ref{overview}. Specifically, taken $N$ reference images $M=\left \{ M_n\in \mathbb{R}^{H\times W}|1\le n\le N  \right \} $ and their corresponding camera position $\{T_n\}$ as input, a two-layer convolutional network is used to calculate their pixel aligned image features $F=\left \{ F_n\in \mathbb{R}^{H\times W\times D}|1\le n\le N  \right \}$ without down sampling. Feature dimension $D$ is set as 128 in this work, and $H,W$ indicates the height and width of an image respectively. The use of multiple reference images provides better multi-view informations. For a 3D query point $p=(x, y, z)\in \mathcal{P} $, we project it back to the 2D image spaces of these references using intrinsics $\{K_n\}$ and camera poses $\{R_n, T_n\}$ and get the corresponding 2D coordinate. Using $p_{n}^{ref}=(u_n, v_n)$ to denote the 2D coordinate in the $n$-th reference image, this projection can be formulated as:
\begin{equation} 
p_{n}^{ref}=\mathcal{M}(p, K_n, R_n, T_n),
\label{projection}
\end{equation}
where $\mathcal{M}$ is the traditional mapping from world space to image space. These corresponding pixel-level features $\{F_{n}(u_n,v_n)\}\in \mathbb{R}^{N\times D}$ from $N$ references are then sampled after a rounding operation and fused with an attention-based module~\cite{locatello2020object} to get the final feature $\tilde{F} = Aggregation(\{F_{n}(u_n,v_n)\})\in \mathbb{R}^{D}$. These feature grids contain rich information about identity and appearance. Using them as an additional condition for our facial radiance field makes the model possible to quickly generalize to a new face appearance from a few observed frames. This dual-driven facial radiance field can be finally formulated as:
\begin{equation} 
\left ( c, \sigma  \right ) = \mathcal{F}_{\theta }\left ( {p}, {d}, {A}, \tilde{F} \right ).
\label{equation2}
\end{equation}

\subsection{Differentiable Face Warping}
\label{3_3}
In Section~\ref{3_2}, we project the query 3D point back to the 2D image spaces of these reference images as Eq.~(\ref{projection}) to get the conditioned pixel features. This operation bases on the prior knowledge in NeRF that intersecting rays casting from different viewpoints should correspond to the same physical location and thus yield the same color~\cite{park2020deformable}. This strict spatial mapping relationship holds for rigid scenes yet the talking face is dynamic. When speaking, the lip and other facial muscles moves according to the pronunciation. Applying Eq.~(\ref{projection}) directly 
on a deformable talking face may result in the key points mismatch. For example, a 3D point near the corner of the mouth in the standard volume space is mapped back to the pixel space of a reference image. If the reference face shows a different mouth shape, the mapped point may fall away from the desired real mouth corner. Such inaccurate mapping results in incorrect pixel feature conditions from reference images, which further affects the prediction of deformations of talking mouth.

To tackle this limitation, we propose an audio-conditioned and 3D point-wise face warping module $\mathcal{D_\eta }$. It regresses offsets $\Delta o=(\Delta  u, \Delta  v)$ for every projected point $p^{ref}$ under the specific deformations, just as shown in the image stream of Fig.~\ref{overview}. Specifically, $\mathcal{D}_\eta$ is realized as a deformation field with a three-layer MLP, where $\eta$ is the learnable parameters. To regress the offset $\Delta o$, dynamics differences between the query image and these reference images need to be effectively exploited. The audio information $A$ reflects the dynamics of the query image, while the deformations of the reference images can be seen through image features $\{F_n\}$ implicitly. We therefore take these two parts together with the query 3D point coordinate $p$ as the input for $\mathcal{D}_\eta$. The process to predict the offset with the face warping module $\mathcal{D}_\eta$ can be formulated as:
\begin{equation} 
\Delta o_n=\mathcal{D}_\eta (p, A, F_{n}(u_n,v_n)).
\label{tt}
\end{equation}
The predicted offset $o_n$ is then added to the $p_{n}^{ref}$ as shown in Fig.~\ref{warp} to get the exact corresponding coordinate ${p_{n}^{ref}}'$ for the 3D query point $p$,
\begin{equation} 
{p_{n}^{ref}}'= {p_{n}^{ref}} + \Delta o_n = (u_n',v_n'),
\label{deform}
\end{equation}
where $u_n'=u_n+\Delta u_n $ and $v_n'=v_n+\Delta v_n $.

Since the hard index operation $F_{n}({u_n}',{v_n}')$ is not differentiable, the gradient cannot be back propagated to this warpping module. We therefore introduce a soft index function to realize the differentiable warpping, where the feature of each pixel is obtained through features interpolation of its surrounding points by bilinear sampling. In this way, the deformation field $\mathcal{D}_\eta$ and the facial radiance field $\mathcal{F}_{\theta }$ can be jointly optimized end to end.
A visualization of this soft index operation is shown in Fig.~\ref{warp}. For the green point, its pixel feature is computed through the features of its four nearest neighbours by bilinear interpolation. To better constrain the training process of this warping module, we introduce a regularization term $L_r$ to limit the value of predicted offsets in a reasonable range to prevent distortions,
\begin{equation} 
L_r = \frac{1}{N\cdot \left |\mathcal{P}  \right | } \sum_{p\in \mathcal{P}}^{} \sum_{n=1}^{N} \sqrt{\Delta u_{n}^2+\Delta v_{n}^2},
\label{tt2}
\end{equation}
where $\mathcal{P}$ is the collection of all 3D points in the voxel space, and $N$ is the number of reference images. Furthermore, we argue that the points with low density are more likely to be background areas that should have low deformation offset. In these regions, stronger regularization constraints should be imposed. For more reasonable constraint, we change the above $L_r$ as:
\begin{equation} 
{L_r}' = (1-\sigma)\cdot L_r,
\label{regulation}
\end{equation}
where $\sigma$ indicates the density of these points. The dynamic facial radiance field can finally be formulated as:
\begin{equation} 
\left ( c, \sigma  \right ) = \mathcal{F}_{\theta }\left ( {p}, {d}, {A}, \tilde{F}' \right ),
\label{equation7}
\end{equation}
where $\tilde{F}' = Aggregation(\{F_{n}({u_n}',{v_n}')\})$.

With this face warping module, all reference images can be transformed to the query space for better modeling the talking face deformations. The ablation study in Section~\ref{ablation} has proven the effectiveness of this component in producing more accurate and audio-synchronized mouth movements.

\subsection{Volume Rendering}
\label{3_4}
The volume rendering is used to integrate the colors $c$ and densities $\sigma$ from Eq.~(\ref{equation7}) into face images. We treat the background, torso and neck parts together as the rendering `background’ and restore it frame by frame from the original videos. We set the color of the last point of each ray as the corresponding background pixel to render a natural background including the torso part. Here we follow the setting in the original NeRF, and the accumulated color $C$ of a camera ray $r$ under the condition of audio signal $A$ and image features $\tilde{F}'$ is:
\begin{equation}
C\left ( r;\theta ,\eta, R, T, A, \tilde{F}' \right )=\int_{z_{near}}^{z_{far}}\sigma\left ( t\right )\cdot c(t)\cdot T\left ( t \right )dt,
\label{nerf}
\end{equation}
where $\theta$ and $\eta$ are the learnable parameters for the facial radiance field $\mathcal{F}_{\theta }$ and the face warping module $\mathcal{D}_\eta$ respectively. $R$ is the rotation matrix and $T$ is the translation vector. $T\left ( t \right )=exp\left (- \int_{z_{near}}^{t} \sigma \left ( r\left ( s \right ) \right )ds\right )$ is the integral transmittance along camera ray, where $z_{near}$ and $z_{far}$ are the near and far bound of the camera ray. We follow the NeRF to design a MSE loss as $L_{MSE}=\left \| C - I \right \|^2$, where $I$ is the ground truth color. Coupled with the regularization term in Eq.~(\ref{regulation}), the overall loss function can be formulated as:
\begin{equation} 
L = L_{MSE} + \lambda \cdot {L_r}'.
\label{loss_all}
\end{equation}

\subsection{Implementation Details}
\label{3_5}
We train only one base radiance field across different identities from coarse to fine. In the coarse training stage, the facial radiance field $\mathcal{F}_\theta$ as Eq.~(\ref{equation2}) is trained under the supervision of $L_{MSE}$ to grasp the structure of the head and establish a general mapping from audio to lip motions. Then we add the face warping module into training as Eq.~(\ref{equation7}) to jointly optimize the offset regression network $\mathcal{D}_\eta$ and the $\mathcal{F}_\theta$ end to end with the loss function $L$ in Eq.~(\ref{loss_all}).

For an arbitrary unseen identity with only a short training clip available, we only need tens of seconds of his/her speaking video for fine-tuning based on the well-trained base model. After short iterations of fine-tuning, the personalized mouth pronunciation patterns can be learned, and the rendered image quality is greatly improved. Then this fine-tuned model can be used for inference.

\section{Experiments}
\subsection{Experimental Settings}

\textbf{Dataset.} AD-NeRF~\cite{guo2021ad} collects several high-resolution videos in natural scenes to better evaluate the performance in practical application. Following this practice, we collect 12 public videos with an average length of 3 minutes from 11 identities from the YouTube. The protagonists of these videos are all celebrities like news anchors, entrepreneurs or presidents. We resample all videos to 25 FPS and set the resolution as $512\times512$. We select three videos from different races and languages (English and Chinese), and combine them into a three-minute video to train the base model. For other videos, we split each of them into three training sets of the length of 10s, 15s and 20s. Then the remaining part is used as the test set. There is no overlap between the training set and the test set. All videos and the corresponding identities used in the following experiments are unseen when training the base model. These data will be released for reproduction. 

\textbf{Head Pose.} Following the AD-NeRF, we estimate head poses based on Face2Face~\cite{thies2016face2face}. To get temporally smooth poses, we further apply the bundle adjustment~\cite{andrew2001multiple} as a temporal filtering. The camera poses $\{R_n, T_n\}$ are the inverse of head poses, where $R$ is the rotation matrix and $T$ is the translation vector.

\textbf{Metrics.} We conduct performance evaluations through some quantitative metrics and visual results. Peak Signal-to-Noise Ratio (PSNR$\uparrow$), Structure SIMilarity (SSIM$\uparrow$)~\cite{wang2004image} and Learned Perceptual Image Patch Similarity (LPIPS$\downarrow$)~\cite{zhang2018unreasonable} are used as image quality metrics. PSNR tends to give higher scores to blurry images~\cite{park2020deformable}. We therefore recommend the more representative perceptual metrics LPIPS. We further use the SyncNet (offset$\downarrow$/confidence$\uparrow$)~\cite{chung2016out} to measure the audio-visual synchronization. The SyncNet offset is better with smaller absolute value. Here we use the `$\downarrow$' as a brief indication.

\textbf{Training Details.} Our code is based on PyTorch~
\cite{paszke2019pytorch}. All experiments are performed on an RTX 3090. The coefficient $\lambda$ in Eq.~(\ref{loss_all}) is set as 5e-8. We train the base model with an Adam solver~\cite{kingma2014adam} for 300k iterations and then jointly train it with the offset regression network for another 100k iterations.

\subsection{Ablation Study}
\label{ablation}

\textbf{The Number of Reference Images.} In this work, we learn a generic rendering from arbitrary reference face images to talking head with the corresponding appearance (including identity, hairstyle and makeup). Here we perform experiments to investigate the performance gains from various reference face images. We select different numbers of references and fine-tune the base model for 10k iterations on 15s video clip respectively. Quantitative comparisons in Table~\ref{refer_num} show that our method is robust to the number of reference images. According to results, we uniformly use four references in the following experiments.

\begin{table}[tb]
\begin{center}
\centering
\renewcommand\tabcolsep{5pt}
\resizebox{\textwidth}{!}{
\begin{tabular}{c |c c|c c |c c|c c } 
\hline
Reference&  \multicolumn{2}{c|}{1} &\multicolumn{2}{c|}{2}& \multicolumn{2}{c|}{4} & \multicolumn{2}{c}{6} \\
\hline
Metric & PSNR$\uparrow$  & LPIPS$\downarrow$& PSNR$\uparrow$  & LPIPS$\downarrow$& PSNR$\uparrow$  & LPIPS$\downarrow$ & PSNR$\uparrow$  & LPIPS$\downarrow$\\
 \hline\hline
 &31.03  &\bf{0.019}  &31.19 &\bf{0.019} &\bf{31.23}  &\bf{0.019} & \bf{31.23} &0.020 \\
\hline
\end{tabular}}
\end{center}
\caption{Quantitative comparisons with different numbers of reference images.}
\label{refer_num}
\end{table}

\begin{table}[tb]
\begin{center}
\centering
\renewcommand\tabcolsep{4pt}
\resizebox{\textwidth}{!}{%
\begin{tabular}{c|c c c |c c c | c c c } 
\hline
\multirow{2}{*}{Method}& \multicolumn{3}{c|}{NeRF~\cite{mildenhall2020nerf}}  &\multicolumn{3}{c|}{AD-NeRF~\cite{guo2021ad}}&\multicolumn{3}{c}{Ours} \\
 &10s &15s&20s&10s &15s&20s&10s &15s&20s  \\
\hline\hline
PSNR$\uparrow$ & 19.83 &19.77&8.02&31.21&\bf{31.32}&30.90&30.95&30.75&30.96\\
SSIM$\uparrow$ &0.773&0.781&0.003&0.948 &0.947&\bf{0.949}& 0.948&0.947& \bf{0.949}\\
LPIPS$\downarrow$ &0.237&0.239&1.058 &0.039&0.041&0.040&\bf{0.036}&\bf{0.036}&\bf{0.036}\\
SyncNet$\downarrow\uparrow$& -&-&-&15/1.313&-14/0.654 & -5/0.932&0/3.447&0/4.105&\bf{0/4.346}\\
\hline
\end{tabular}}
\end{center}
\caption{Method comparisons when using different lengths of training videos.}
\label{diflength}
\end{table}

\begin{table}[t]
\begin{center}
\centering
\renewcommand\tabcolsep{5pt}
\resizebox{\textwidth}{!}{%
\begin{tabular}{c|c c c c |c c c c } 
\hline
\multirow{2}{*}{Method} & \multicolumn{4}{c|}{Test Set A}  &\multicolumn{4}{c}{Test Set B} \\
 &PSNR$\uparrow$ &SSIM$\uparrow$ & LPIPS$\downarrow$& SyncNet$\downarrow\uparrow$  & PSNR$\uparrow$ &SSIM$\uparrow$ & LPIPS$\downarrow$& SyncNet$\downarrow\uparrow$ \\
\hline\hline
GT &-&-&-&4/7.762& -&- &-& 3/8.947\\
 w/o& 29.50 &0.907 &0.057 &-1/4.152& 28.98&\bf{0.899}&0.104&-2/2.852 \\
 w &\bf{29.66}&\bf{0.911}&\bf{0.053}&\bf{0/4.822}&\bf{29.14}&\bf{0.899}&\bf{0.101}&\bf{0/4.183}\\
\hline
\end{tabular}}
\end{center}
\caption{Ablation study to investigate the contribution of the proposed differentiable face warping module. `w' indicates the model equipped with the face warping module.}
\label{warp_table}
\end{table}

\begin{figure}[htb]
\begin{center}
\includegraphics[width=1.0\linewidth]{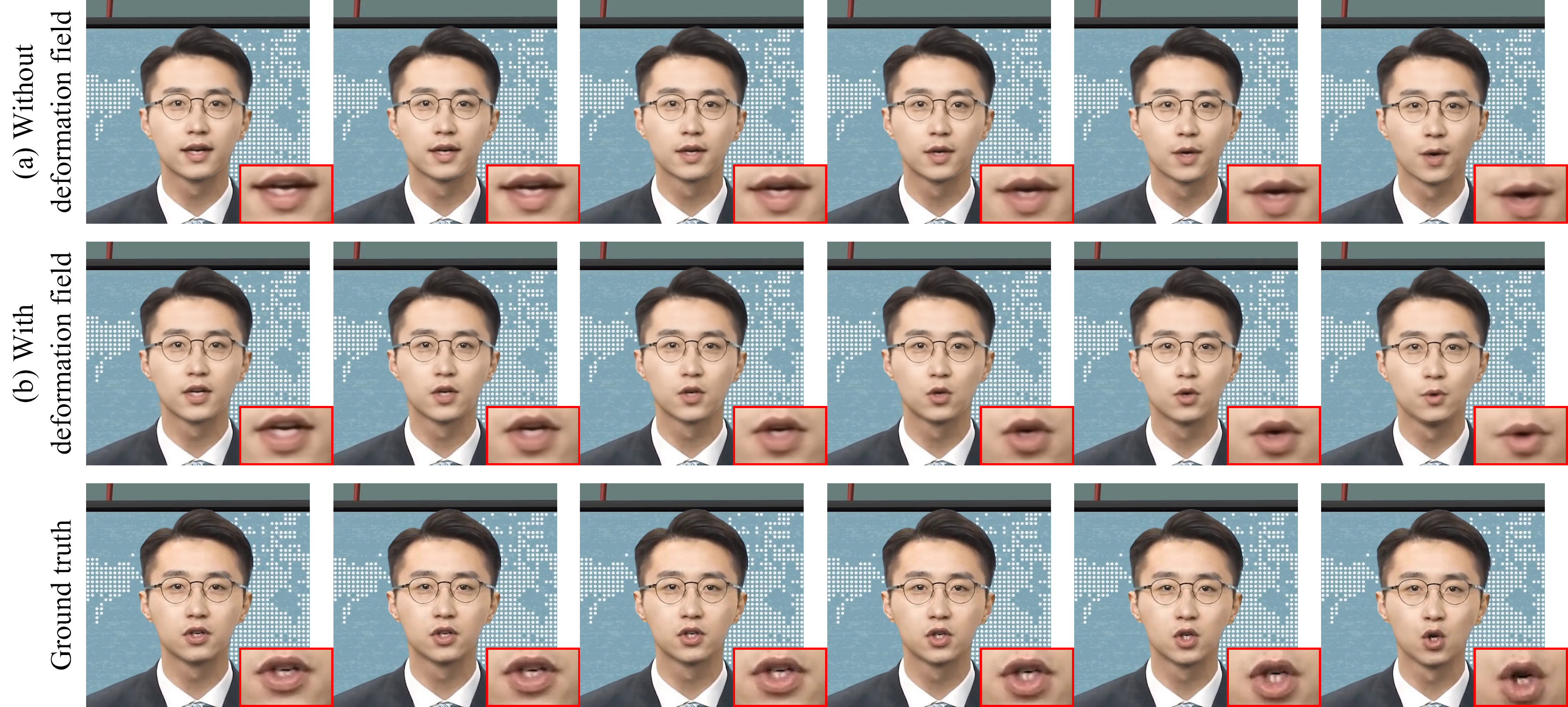}
\end{center}
\caption{Ablation study on the proposed face warping module. The ground truth sequence shows a pout-like expression. Generated results from the model equipped with the deformation field reproduce such pronunciation trend well in line (b), while results in line (a) hardly reflect such lip motions.}
\label{w_wo}
\end{figure}

\textbf{Impact of the Length of Training Data.} In this
subsection, we investigate the impact of different amounts of training data. We fine-tune the proposed DFRF with 10s, 15s, and 20s training videos for 50k iterations. For fair comparisons, we train NeRF and AD-NeRF with the same data and iterations. It is worth noting that we have tried to pre-train NeRF and AD-NeRF across identities following DFRF. However since they lack the ability to generalize between different identities, such per-training fails to learn the general audio-lip mapping. Experimental results in Table~\ref{diflength} show that tens of seconds of data are insufficient for NeRF training. PSNR tends to give higher scores to blurry images~\cite{park2020deformable}, so we recommend LPIPS as more representative metrics for visual quality. In comparison, our method is able to acquire more prior knowledge about the general audio-lip mapping from the base model, thus achieving better audio-visual sync with limited training data. With only a 10s training video, the proposed DFRF can achieve superior 0.036 LPIPS and 3.447 SyncNet confidence, while AD-NeRF struggles in the lip-audio sync.

\begin{table}[ht]
\begin{center}
\centering
\renewcommand\tabcolsep{3.5pt}
\resizebox{\textwidth}{!}{%
\begin{tabular}{c | c|c c c >{\columncolor{red!10}} c |c c c >{\columncolor{red!10}}c } 
\hline
 \multicolumn{2}{c|}{\multirow{2}{*}{Method}}& \multicolumn{4}{c|}{Test Set A}  &\multicolumn{4}{c}{Test Set B} \\
\multicolumn{2}{c|}{}&PSNR$\uparrow$ &SSIM$\uparrow$ & LPIPS$\downarrow$& SyncNet$\downarrow\uparrow$  & PSNR$\uparrow$ &SSIM$\uparrow$ & LPIPS$\downarrow$& SyncNet$\downarrow\uparrow$ \\
\hline\hline
\multicolumn{2}{c|}{Ground-truth} & -& -& -& 0/7.217& -&- &-& -1/7.762\\
\hline
 \multirow{3}{*}{NeRF} & 1k& 16.88 & 0.708 &0.198 & -  &14.69 & 0.397 & 0.442&- \\
 & 10k& 13.98 &0.531 &0.338 & - &15.24 & 0.396 & 0.427 &-\\
 & 40k& 15.87 &0.556 &0.306 &-  &15.91 & 0.405&0.394&-\\
 \hline
\multirow{3}{*}{AD-NeRF} & 1k&27.38&	0.901&0.084&	-15/0.136 &27.61&0.863&0.115&14/0.798 \\
& 10k &29.14&	0.931&0.057&-14/0.467 &30.07&0.905&0.083&-2/0.964 \\
& 40k &29.45&	0.936&	0.039&	-14/0.729 & \bf{30.72}&0.909&0.059 &-2/1.017 \\
\hline
\multirow{3}{*}{Ours} & 1k &28.96&	0.933&	0.040&	-1/2.996&  29.05&0.892&0.076&0/3.157\\
&10k&29.33 &0.935&0.043&	0/4.246
 &29.68&0.905&0.063&0/4.038\\
&40k &\bf{29.48}&\bf{0.937}&\bf{0.037}&\bf{1/4.431} &30.44&\bf{0.925}&\bf{0.045}&\bf{0/4.951}\\
\hline
\end{tabular}}
\end{center}
\caption{Method comparisons on two test sets using 15s training clip for different training iterations. More visual results can be seen in Fig.~\ref{fig5} and Fig.~\ref{all}.}
\label{table4}
\end{table}

\begin{figure}[htb]
\begin{center}
\includegraphics[width=1\linewidth]{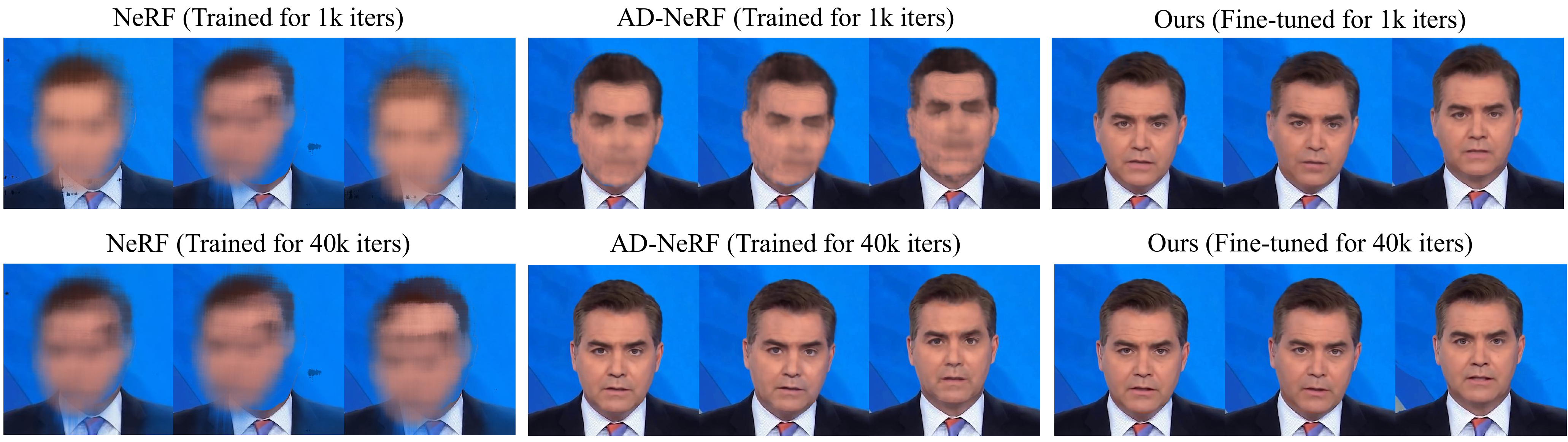}
\end{center}
\caption{Visual comparison using 15s training clip for different training iterations.}
\label{fig5}
\end{figure}

\begin{figure}[htb]
\begin{center}
\includegraphics[width=1\linewidth]{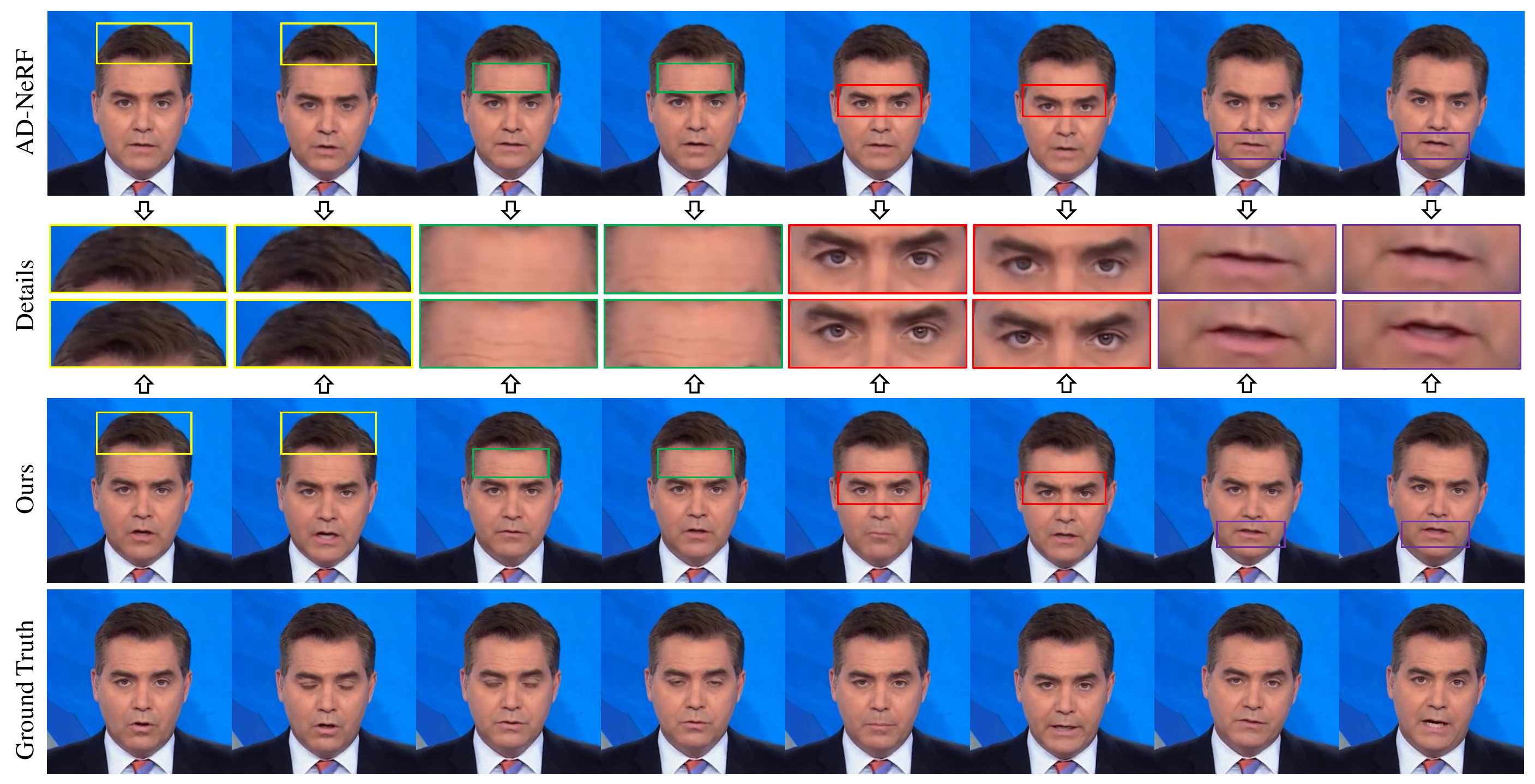}
\end{center}
\caption{Comparison with AD-NeRF using the same 15s training clip for 40k training iterations. We zoom in on some facial details for better visual quality comparison.}
\label{all}
\end{figure}

\textbf{Effect of Differentiable Face Warping.} In DFRF, we propose an audio conditioned differentiable face warping module for better modeling the dynamics of talking face. Here we conduct an ablation study to investigate the contribution of this component. Table~\ref{warp_table} shows the generated results with and without warping module on two test sets. All models are fine-tuned on 15s videos for 50k iterations. Without this module, the query 3D point cannot be mapped to the exact corresponding point in the reference image, especially in some areas with rich dynamics. Therefore, the dynamics of the speaking mouth are affected to some extent, which is reflected in the audio-visual sync (SyncNet score). In contrast, the model equipped with the deformation field can significantly improve the SyncNet confidence and the visual quality also has slight improvement. Fig.~\ref{w_wo} further shows some visual results for more intuitive comparisons. In this video sequence, the ground truth shows a pout-like expression. The generated results (b) with the deformation field show such pronunciation trend well, while results in (a) hardly reflect this kind of lip motions.

\subsection{Method Comparisons}
\textbf{Method Comparisons in the Few-shot Setting.} In this section, we perform method comparisons on two test sets using a 15s training clip for different training iterations. Quantitative results in Table~\ref{table4} show that our proposed method far surpasses NeRF and AD-NeRF in the perceptual image quality metric LPIPS. PSNR tends to give higher scores to blurry images~\cite{park2020deformable} which can be proved in the visualization in Fig.~\ref{all}, so we recommend LPIPS as more representative metrics for visual quality. We also achieve higher audio-lip synchronization indicated by the SyncNet score while AD-NeRF nearly fails on this indicator. Fig.~\ref{fig5} visualizes the generated frames of the three methods. Under the same 1k training iterations, the visual quality of our method is far superior to others. When training for 40k iterations, the AD-NeRF achieves acceptable visual quality, however some face details are missing. The visual gap with our method can be seen obviously from the zoomed-in details in Fig.~\ref{all}. We show two generated talking sequences driven by the same audio from our method and AD-NeRF with 15s training clip after 40k iterations in Fig.~\ref{all}. Compared with the ground truth, our method shows more accurate audio-lip synchronization than AD-NeRF. For example, in the fifth frame, the rendered face from AD-NeRF opens the mouth wrongly. We zoom in some facial details for clearer comparison. It can be seen that our method has generated more realistic details such as sharper hair texture, more obvious wrinkles, brighter pupils and more accurate mouth shape. In our supplementary video, we further add the visual comparison with AD-NeRF when it is trained to convergence (400k iterations).

\textbf{Method Comparisons with More Training Data.} Our DFRF is far superior to others in the few-shot learning setting. For more comprehensive evaluations, we further compare the DFRF with some recent high-performance non-NeRF 3D-based methods \cite{suwajanakorn2017synthesizing,thies2020neural} and the AD-NeRF~\cite{guo2021ad} with more training data (180s training clip). Since the source of \cite{suwajanakorn2017synthesizing,thies2020neural} are not fully open, we follow the AD-NeRF to collect two test sets from the demos of \cite{suwajanakorn2017synthesizing,thies2020neural} for method comparisons, and the results are shown in Table~\ref{moredata}. Our method still surpasses others with long training clip up to 180s, since the proposed face warping module better models the talking face dynamics. Moreover, our DFRF is the only method that works in the few-shot learning setting. In the supplementary video, we further include more comparisons with 2D-based (non-NeRF based) methods.

\begin{table}[tb]
\begin{center}
\centering
\renewcommand\tabcolsep{4.5pt}
\resizebox{\textwidth}{!}{%
\begin{tabular}{c |c c c |c c c |c } 
\hline
\multirow{2}{*}{Method} & \multicolumn{3}{c|}{Test Set A} &\multicolumn{3}{c|}{Test Set B} & Few-shot \\
& PSNR$\uparrow$& LPIPS$\downarrow$ & SyncNet$\downarrow\uparrow$  & PSNR$\uparrow$& LPIPS$\downarrow$ & SyncNet$\downarrow\uparrow$& Method \\
\hline\hline
Suwajanakorn~\emph{et al.}~\cite{suwajanakorn2017synthesizing} & -&-&3/4.301&-&-&- &\ding{53}\\
NVP~\cite{thies2020neural} &-&-&-&-&-&-1/4.677&\ding{53} \\
AD-NeRF~\cite{guo2021ad} & 33.20&0.032&0/5.289&33.85&	0.028&0/4.200&\ding{53} \\
 \hline
Ours&\bf{33.28}&\bf{0.029}&\bf{1/5.301}&\bf{34.65}&\bf{0.027}&\bf{1/5.755}&\checkmark\\
\hline
\end{tabular}
}
\end{center}
\caption{Method comparisons with two non-NeRF based methods SO~\cite{suwajanakorn2017synthesizing} and NVP~\cite{thies2020neural} and the AD-NeRF~\cite{guo2021ad} under the setting with more training data.}
\label{moredata}
\end{table}

\begin{table}[tb]
\renewcommand\tabcolsep{4.5pt}
\begin{center}
\resizebox{\textwidth}{!}{%
 \begin{tabular}{c| c  c  c  c c c c} 
     \hline
     \multirow{2}{*}{Source-Target} & Same & English  & Chinese & Russian & French & Spanish &German \\
     & Identity & (Male) &  (Male) &  (Male) & (Female) & (Female) &(Female) \\
     
     \hline\hline
     English (Male) &-3/5.042&	-2/3.805&	-2/4.879&	-1/4.118&	-2/3.019&	-2/4.986 &-1/4.820\\ 
     Chinese (Male) &0/4.486&	-2/3.029&	-2/3.534&	-3/4.206&	-3/3.931&	-3/4.085 &-2/4.494\\ 
     Russian (Male) &-1/4.431&	-2/2.831&	-2/4.397&	-2/5.109&	-3/4.307&	-1/5.011 &-1/5.008\\ 
     French (Male) & -2/4.132&	-2/3.193&	-3/3.383&	-3/4.088&	-2/3.339&	-2/3.728&	-1/3.529\\ 
     \hline
 \end{tabular}
 }
 \end{center}
 \caption{SyncNet scores under the cross language setting.}
 \label{crosslan}
\end{table}

\textbf{Cross-Language Results.} 
We further verify the performance of our method driven by audios with different languages and genders. We select four models trained with 15s training clips from different languages (source), then conduct inference with driven audios cross six languages and different genders (target). We also list the self-driven (source and target are from the same identity) results (the second column) for reference. SyncNet (offset/ confidence)($\downarrow$ / $\uparrow$) scores in Table~\ref{crosslan} shows that our method produces reasonable lip-audio synchronization in such cross language setting.

\subsection{Applications and Ethical Considerations}
The talking head synthesis technique can be used in a variety of practical scenarios, including correcting pronunciation, re-dubbing, virtual avatars, online education, electronic game making and providing speech comprehension for hearing impaired people. However, the talking head technology may bring some potential misuse issues. We are committed to combating these malicious behaviors and advocate more attention to the active application of this technology. We support those organizations that devote themselves to identifying fake defamatory videos, and are willing to provide them with the generated videos to expand the training set for automatic identification technology. Meanwhile, any individual or organization should obtain our permission before using our code, and it is recommended to use a watermark to indicate the generated video.

\section{Conclusion}
In this paper, we have proposed a dynamic facial radiance field for few-shot talking head synthesis. We employ audio signals coupled with 3D-aware image features as the condition for fast generalizing to novel identities. To better model the mouth motions of talking head, we further learn an audio-conditioned face warping module to deform all reference images to the query space. Extensive experiments show the superiority of our method in generating natural talking videos with limited training data and iterations.

\subsubsection{Acknowledgments.}
This work was supported in part by the National Key Research and Development Program of China under Grant 2017YFA0700802, in part by the National Natural Science Foundation of China under Grant 62125603 and Grant U1813218, in part by a grant from the Beijing Academy of Artificial Intelligence (BAAI).

\clearpage
%
%
\bibliographystyle{splncs04}
\bibliography{egbib}
\end{document}